\documentclass[conference]{IEEEtran}
\IEEEoverridecommandlockouts

\usepackage{cite}
\usepackage{amsmath,amssymb,amsfonts}
\usepackage{algorithmic}
\usepackage{graphicx}
\usepackage{textcomp}
\usepackage{xcolor}
\def\BibTeX{{\rm B\kern-.05em{\sc i\kern-.025em b}\kern-.08em
    T\kern-.1667em\lower.7ex\hbox{E}\kern-.125emX}}

\usepackage{todonotes}
\usepackage[colorlinks=true, allcolors=blue]{hyperref}
\usepackage{booktabs}
\usepackage{multirow}
\usepackage{makecell}
\usepackage{tipa}
\usepackage{subcaption}
\usepackage{tikz, pgfplots}
\usetikzlibrary{positioning}
\usepgfplotslibrary{fillbetween}
\pgfplotsset{compat=newest}
\usepackage{comment}

\newcommand{\etal}{et al.}

\newcommand{\eg}{e.g.}
\newcommand{\quotes}[1]{``#1''}

\newif\ifshowtodos
\showtodosfalse

\newenvironment{customlegend}[1][]{%
    \begingroup
    \csname pgfplots@init@cleared@structures\endcsname
    \pgfplotsset{#1}
}{
    \csname pgfplots@createlegend\endcsname
    \endgroup
}

\def\addlegendimage{\csname pgfplots@addlegendimage\endcsname}

\pgfplotsset{
    /pgfplots/xlabel near ticks/.style={
        /pgfplots/every axis x label/.style={
            at={(ticklabel cs:0.5)},anchor=west ticklabel
        }
    },
    /pgfplots/ylabel near ticks/.style={
        /pgfplots/every axis y label/.style={
            at={(ticklabel cs:0.5)},rotate=90,anchor=west ticklabel
        }
    }
}

\definecolor{PINK}{HTML}{FF0055}
\definecolor{BLUE}{HTML}{008BE7}
\definecolor{RED}{HTML}{FF0000}
\definecolor{LIMEGREEN}{HTML}{32CD32}

\definecolor{LA}{HTML}{0173B2}
\definecolor{TTCD}{HTML}{DE8F05}
\definecolor{TBCD}{HTML}{029E73}
\definecolor{VEL}{HTML}{D55E00}

\definecolor{S1}{HTML}{FF0000}
\definecolor{S2}{HTML}{FFBD00}
\definecolor{S3}{HTML}{83FF00}
\definecolor{S4}{HTML}{00FF39}
\definecolor{S5}{HTML}{00FFF5}
\definecolor{S6}{HTML}{004BFF}
\definecolor{S7}{HTML}{7100FF}
\definecolor{S8}{HTML}{FF00CF}

\definecolor{arytenoid_cartilage}{HTML}{8A2BE2}
\definecolor{epiglottis}{HTML}{40E0D0}
\definecolor{lower_incisor}{HTML}{00FFFF}
\definecolor{lower_lip}{HTML}{00FF00}
\definecolor{pharynx}{HTML}{DAA520}
\definecolor{soft_palate}{HTML}{1E90FF}
\definecolor{soft_palate_midline}{HTML}{1E90FF}
\definecolor{thyroid_cartilage}{HTML}{F4A460}
\definecolor{tongue}{HTML}{FF8C00}
\definecolor{upper_incisor}{HTML}{FFFF00}
\definecolor{upper_lip}{HTML}{FF00FF}
\definecolor{vocal_folds}{HTML}{FF69B4}

\definecolor{embedding0}{HTML}{FF0000}
\definecolor{embedding1}{HTML}{FFD500}
\definecolor{embedding2}{HTML}{4FFF00}
\definecolor{embedding3}{HTML}{00FF86}
\definecolor{embedding4}{HTML}{009EFF}
\definecolor{embedding5}{HTML}{3700FF}
\definecolor{embedding6}{HTML}{FF00ED}

\begin{document}

\title{Evaluating Speech Articulation Synthesis with Articulatory Phoneme Recognition}

\author{\IEEEauthorblockN{\textbf{Vinicius Ribeiro}$^{\dagger}$ \\}
\IEEEauthorblockA{\textit{Université de Lorraine, CNRS, Inria, LORIA} \\
F-54000, Nancy, France \\
0000-0001-5897-5765 \\}
\and
\IEEEauthorblockN{\textbf{Yves Laprie} \\}
\IEEEauthorblockA{\textit{Université de Lorraine, CNRS, Inria, LORIA} \\
F-54000, Nancy, France \\
0000-0002-2379-6481 \\}
}

\maketitle

\begin{tikzpicture}[remember picture, overlay]
\node[anchor=south, font=\small\itshape] at (current page.south) [yshift=1cm] {Accepted for publication at the European Signal Processing Conference (EUSIPCO), 2026.};
\end{tikzpicture}

{\renewcommand{\thefootnote}{$\dagger$}\footnotetext{This work was conducted while the author was a PhD student at the affiliated laboratory.}}

\begin{abstract}

Recent advances in machine learning and the availability of articulatory datasets allow vocal tract synthesis to be conditioned on phonetic sequences, a primary task of articulatory speech synthesis. However, quality assessment needs a better definition. Generally, ranking generative models is tricky due to subjectivity. However, articulatory synthesis has the additional difficulty of requiring specialized knowledge in vocal tract anatomy and acoustics. To address this problem, this paper proposes to evaluate speech articulation synthesis using phoneme recognition as a proxy.

Our hypothesis is that phoneme recognition using articulatory features better captures nuances in phoneme production, such as correct places of articulation, which traditional metrics (\eg, point-wise distance metrics) do not. We train a neural network with acoustic and articulatory features extracted from a single-speaker RT-MRI dataset. Then, we compare the recognition performance when testing the model with different synthetic articulatory features. Our results show that our articulatory feature set is phonetically rich and helps exploring additional dimensions on speech articulation synthesis.

\end{abstract}

\section{Introduction}

Ribeiro~\etal~\cite{ribeiro2023deep} described how to synthesize the shape of the vocal tract conditioned in the sequence of phonemes to be articulated, exploring several methods for synthesizing vocal tract articulators during speech. The baseline method was a phoneme-wise mean-contour, which computes the average contour for each phoneme in a single-speaker real-time MRI (RT-MRI) dataset. Then model-free~\cite{ribeiro2021towards,RIBEIRO2022} and autoencoder-based~\cite{ribeiro2022autoencoder} vocal tract shape synthesizers were used to generate complete vocal tract shapes from the sequences of phonemes to be articulated. On the one hand, the two latter models consistently outperformed the baseline and presented indistinguishable performances between each other in terms of point-to-closest-point distance. On the other hand, analysis of tract variables indicated that the autoencoder-based method learns better places of articulation, performing better constrictions than the model-free system. However, assessing the quality of synthesized articulatory features from speech remains a challenge.

Point-wise distance metrics are easy to interpret, but the usage is limited due to substantial inter- and intra-speaker variability. In contrast, measuring tract variables associated with each target phoneme fits well consonants but does not suit vowels because the latter is characterized by the resonator's shape and not by constrictions. A better metric for vowels would be the measurement of formant frequencies~\cite{serrurier2022f1} by solving simplified aero-acoustic equations in the synthetic vocal tract, 
which is computationally expensive.
Nevertheless, the two metrics were inconclusive regarding the two proposed models, being satisfactory only for comparing with the phoneme-wise mean-contour, which is a very simplistic model. Subjective analysis indicated that synthetic utterances using the model-free approach seem more stable and temporally consistent than those of the autoencoder-based system, but the traditional metrics implemented fail to reflect this perception.

Recent research has paid significant attention to the classification of articulatory characteristics and their use in understanding the relationship between articulations and acoustics and how neural networks map the two. Elie~ \etal~\cite{elie2023optimal} used phoneme recognition probability as a measure of intelligibility in their cost function. Saha~\etal~\cite{saha2018towards} trained a Long-Term Recurrent Convolutional Network to classify 51 vowel-consonant-vowel (VCV) contexts from RT-MRI films from 17 speakers, obtaining an accuracy of $42\%$. Van~Leeuwen~\etal~\cite{van2019cnn} trained a CNN to classify sustained phonemes (vowels and fricatives) from static mid-sagittal MRI and obtained an accuracy of $57\%$. Interestingly, the model learned representations compatible with the vowel chart, showing that although accuracy is limited, the model is consistent with standard phonetic knowledge. In the problem of evaluating synthesized vocal tract shapes, Engwall~\cite{engwall2006evaluation} used an articulatory classifier as an evaluation metric for acoustic-to-articulatory inversion of VCV words in Swedish sentences using linear estimation and neural networks. Engwall's research shows that the articulatory classifier provides a more intelligible metric than the RMS error and correlation coefficients. Inspired by these articles, we took a similar direction by using phoneme recognition to measure phonetic information in mid-sagittal RT-MRI contours. We analyzed speech articulations generated by vocal tract shape synthesizers from the literature using phoneme recognition. We first trained a phoneme recognizer on the acoustic signal as a baseline. Then, we trained the recognizer on the real articulators' contours (true articulatory features). Since the mid-sagittal RT-MRI does not include vocal fold excitation, we add a categorical encoding representing voicing information.

We quantify the information retained by the vocal tract contours by comparing the recognition error with the acoustic signal and the true articulatory features with and without voicing encoding. Next, the vocal tract shapes of the utterances in the test set were synthesized using the phoneme-wise mean-contour, model-free and autoencoder-based systems. These synthetic features with voicing encoding are input into the phoneme recognizer trained with true articulations. The recognition error of this test exhibits how much phonetic information the synthesizer can reproduce. We hypothesize that if the true contours carry enough information, the synthetic articulations should also exhibit a recognition performance comparable to the true articulatory features.

\section{Dataset}

\begin{figure}
    \centering
    \begin{tabular}{cc}
    \includegraphics[width=0.5\linewidth,trim={3cm 3cm 3cm 3cm},clip]{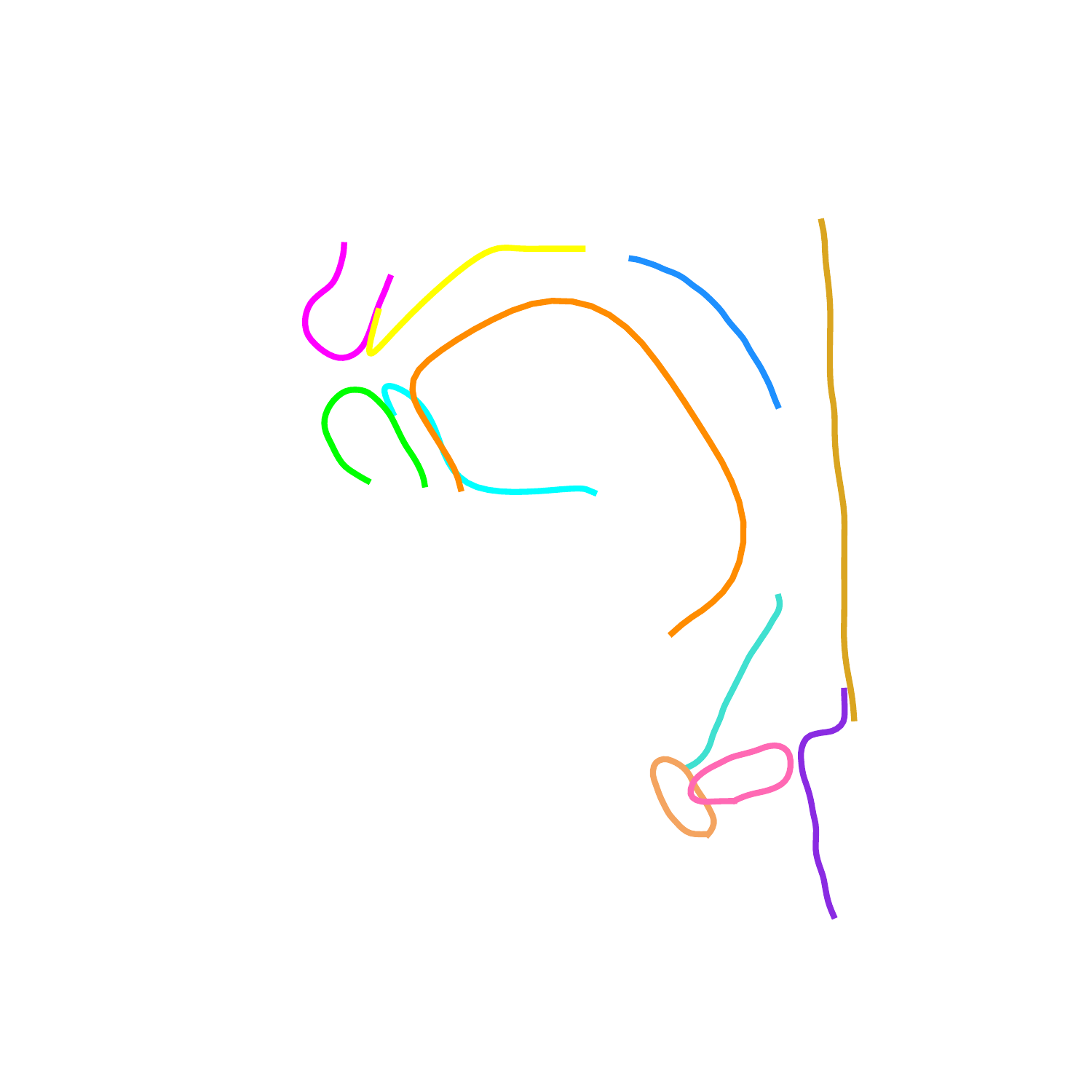} &
    \begin{tikzpicture}[scale=1.0]
         \scriptsize
         \begin{customlegend}[legend entries={
            Arytenoid Cart.,
            Epiglottis,
            Lower Incisor,
            Lower Lip,
            Pharynx,
            Soft Palate,
            Thyroid Cart.,
            Tongue,
            Upper Incisor,
            Upper Lip,
            Vocal Folds
        }, legend columns=1, legend cell align={left}]
             \addlegendimage{-,line width=1.5pt,arytenoid_cartilage}
             \addlegendimage{-,line width=1.5pt,epiglottis}
             \addlegendimage{-,line width=1.5pt,lower_incisor}
             \addlegendimage{-,line width=1.5pt,lower_lip}
             \addlegendimage{-,line width=1.5pt,pharynx}
             \addlegendimage{-,line width=1.5pt,soft_palate_midline}
             \addlegendimage{-,line width=1.5pt,thyroid_cartilage}
             \addlegendimage{-,line width=1.5pt,tongue}
             \addlegendimage{-,line width=1.5pt,upper_incisor}
             \addlegendimage{-,line width=1.5pt,upper_lip}
             \addlegendimage{-,line width=1.5pt,vocal_folds}
         \end{customlegend}
     \end{tikzpicture} \\
    \end{tabular}
    \caption[Articulatory features used for phoneme recognition.]{Articulatory features used for phoneme recognition plus the upper incisor, which is the reference for the coordinate system.}
    \label{fig:articulatory_features}
\end{figure}

Our corpus contains 2.5 hours of speech from a single female native French speaker. It is composed of RT-MRI images (50 Hz), de-noised audio signals and manually corrected phonetic annotations. To our knowledge, this database is the largest dynamic MRI database for a speaker, which ensures the relevance of the training process. The acoustic features were obtained by computing Mel spectrograms with 80 frequency bands. The contours of ten vocal tract articulators were extracted from the images using the method described in Ribeiro~\etal~\cite{ribeiro2024automatic}. These contours are also used to learn the link between the acoustic signal and the shape of the vocal tract as part of our work on articulatory to acoustic inversion~\cite{azzouz:hal-05293831}. The articulators are the arytenoid cartilage, epiglottis, lower incisor, lower lip, pharynx, soft palate, thyroid cartilage, tongue, upper lip, and vocal folds. In addition, the upper incisor is used as a reference for the coordinate system. However, it is kept apart from the experiments. The contours of the ten individual articulators ($x_i \in \mathbb{R}^{2\times50},~i \in [1, 10]$) from \autoref{fig:articulatory_features}, where $50$ refers to the number of samples in each curve, were concatenated composing a 2-channel articulatory feature vector with the x- and y-axis in the channel dimension ($x^\prime \in \mathbb{R}^{2\times500}$). We trained the vocal tract shape synthesizer from \cite{RIBEIRO2022, ribeiro2022autoencoder} with this work's dataset to obtain the synthetic articulatory features.

\autoref{tab:dataset_desc} summarizes the number of utterances and duration of each dataset split. The phonetic vocabulary comprises $50$ tokens, from which $42$ are phonetic tokens, and $8$ are non-phonetic, representing blank token, silence, unknown token, and noise after /\textipa{i}, \textipa{e}, \textipa{u}, \textipa{y}, \textipa{\o}/. Unvoiced plosives are characterized by two phases: closure and burst. Therefore, the phonemes /\textipa{p}, \textipa{t}, \textipa{k}/ are represented by two tokens, one for each phase. Closure and burst phases are less easy to detect for voiced plosives, so we decided not to segment them. Phonemes were grouped according to their places of articulation for the evaluation as seen in \autoref{tab:phonetic_classes} -- phonemes not present in the table were classified as \quotes{others}.

\begin{table}[hbt!]
    \centering
    \caption{Summary of the train, validation and test splits.}
    \begin{tabular}{lcc}
    \toprule
    \thead{Dataset} & \thead{Number of Utterances} & \thead{Duration (Minutes)} \\
    \midrule
    Train & 1\,399 & 125.1 \\
    Validation & 116 & 11.3 \\
    Test & 114 & 11.2 \\
    \midrule
    Total & 1\,629 & 147.6 \\
    \bottomrule
    \end{tabular}
    \label{tab:dataset_desc}
\end{table}

\begin{table}
    \centering
    \caption[Phonemes considered under each phonetic class.]{Phonemes considered under each phonetic class. Phonemes with similar places of articulation are put grouped together.}
    \begin{tabular}{ll}
        \toprule
        \thead{Phonetic Classes} & \thead{Phonemes} \\
        \midrule
        Dental & \textipa{t}, \textipa{d}, \textipa{n}, \textipa{l}, \textipa{z}, \textipa{s}\\
        \midrule
        Labial & \textipa{p}, \textipa{b}, \textipa{m}, \textipa{f}, \textipa{v} \\
        \midrule
        Palatal & \textipa{k}, \textipa{g}, \textipa{Z}, \textipa{S}, \\
        \midrule
        Front Vowels & \textipa{i}, \textipa{e}, \textipa{E}, \textipa{\~E}/\textipa{\~\oe}, \textipa{j} \\
        \midrule
        Back Vowels & \textipa{u}, \textipa{o}, \textipa{O}, \textipa{\~o}, \textipa{w} \\
        \midrule
        Open Vowels & \textipa{a}, \textipa{\~a} \\
        \midrule
        Front Rounded Vowels & \textipa{y}, \textipa{\o}, \textipa{\oe}, \textipa{4} \\
        \bottomrule
    \end{tabular}
    \label{tab:phonetic_classes}
\end{table}

\section{Methods}

The contours of the arytenoid cartilage, epiglottis center line, lower incisor, lower lip, pharynx, soft palate center line, thyroid cartilage, tongue, upper lip, and vocal folds were concatenated to compose the articulatory features. The $x$- and $y$-coordinates form a 2-channel, 500-dimensional feature vector ($10~\mathrm{articulators} \times 50~\mathrm{samples~per~curve}$). The synthetic articulatory features are obtained by inputting the test utterances into the synthesizers presented in \cite{ribeiro2023deep}, which return the synthetic articulatory features. The phonemes were grouped by their places of articulation for the evaluation (see \autoref{tab:phonetic_classes}).

\begin{figure}
    \centering
    \begin{tabular}{cc}
    \subcaptionbox{Adapter block\label{sfig:phoneme_recognizer_arch-a}}
    {\includegraphics[width=0.28\linewidth]{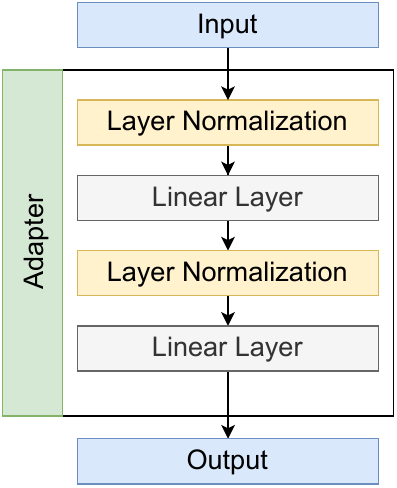}} &
    \subcaptionbox{Phoneme recognizer architecture with one residual CNN block and one recurrent block\label{sfig:phoneme_recognizer_arch-b}}
    {\includegraphics[width=0.57\linewidth]{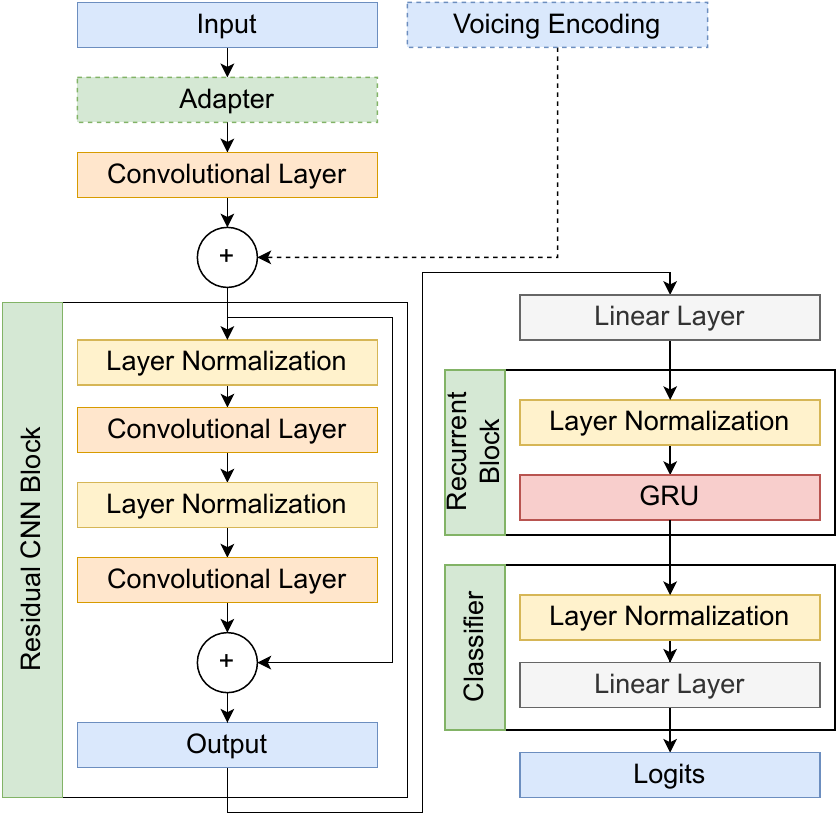}}
    \end{tabular}
    \caption[Phoneme recognition network architecture.]{Phoneme recognition network architecture.}
    \label{fig:phoneme_recognizer_arch}
\end{figure}

The Deep Speech 2~\cite{amodei2016deep} architecture inspires the phoneme recognizer.
The network comprises convolutional blocks with a residual additive connection between the inputs and the outputs, followed by recurrent blocks. Finally, a block of linear layers composes the classifier.
To fit the articulatory features into the model, we prepend to the initial convolutional layer an adapter block formed by linear layers that convert the 500-dimensional tensor into an 80-dimensional feature vector. When voicing encoding was used, it was added to the outputs of the first convolutional layer. \autoref{fig:phoneme_recognizer_arch} presents a schematic of the network architecture. Our implementation uses five residual convolutional blocks and three recurrent blocks.

The CTC loss~\cite{graves2006connectionist} was used as the learning objective, and the phoneme error rate (PER), measured in terms of the Levenshtein distance~\cite{levenshtein1966binary}, is the evaluation metric. Furthermore, we computed the t-Distributed Stochastic Neighbor Embedding (t-SNE)~\cite{van2008visualizing} representations of the models' features calculated immediately before the classifier layer. The network was trained using the Adam optimizer~\cite{kingma2014adam} and the cyclic learning rate scheduler policy~\cite{smith2017cyclical}. Additionally, we apply a slight Gaussian noise to the logits (model's outputs before the softmax) as a regularization strategy together with $L_2$ regularization.

The code is publicly available at our repository on Github\footnote{https://github.com/vribeiro1/artspeech}.

\section{Results}

\begin{figure*}[h!]
    \centering
    \begin{tabular}{cccc}
         \multicolumn{4}{c}{\begin{tikzpicture}[scale=1.0]
             \scriptsize
             \begin{customlegend}[legend entries={
                Dental,
                Labial,
                Palatal,
                Front Vowels,
                Back Vowels,
                Open Vowels,
                Front Rounded Vowels
            }, legend columns=7, legend cell align={left}]
                 \addlegendimage{only marks,mark=*,embedding0}
                 \addlegendimage{only marks,mark=*,embedding1}
                 \addlegendimage{only marks,mark=*,embedding2}
                 \addlegendimage{only marks,mark=*,embedding3}
                 \addlegendimage{only marks,mark=*,embedding4}
                 \addlegendimage{only marks,mark=*,embedding5}
                 \addlegendimage{only marks,mark=*,embedding6}
             \end{customlegend}
         \end{tikzpicture}} \\
         \subcaptionbox{Acoustic signal\label{sfig:phoneme_representations-a}}{\includegraphics[width=0.225\linewidth]{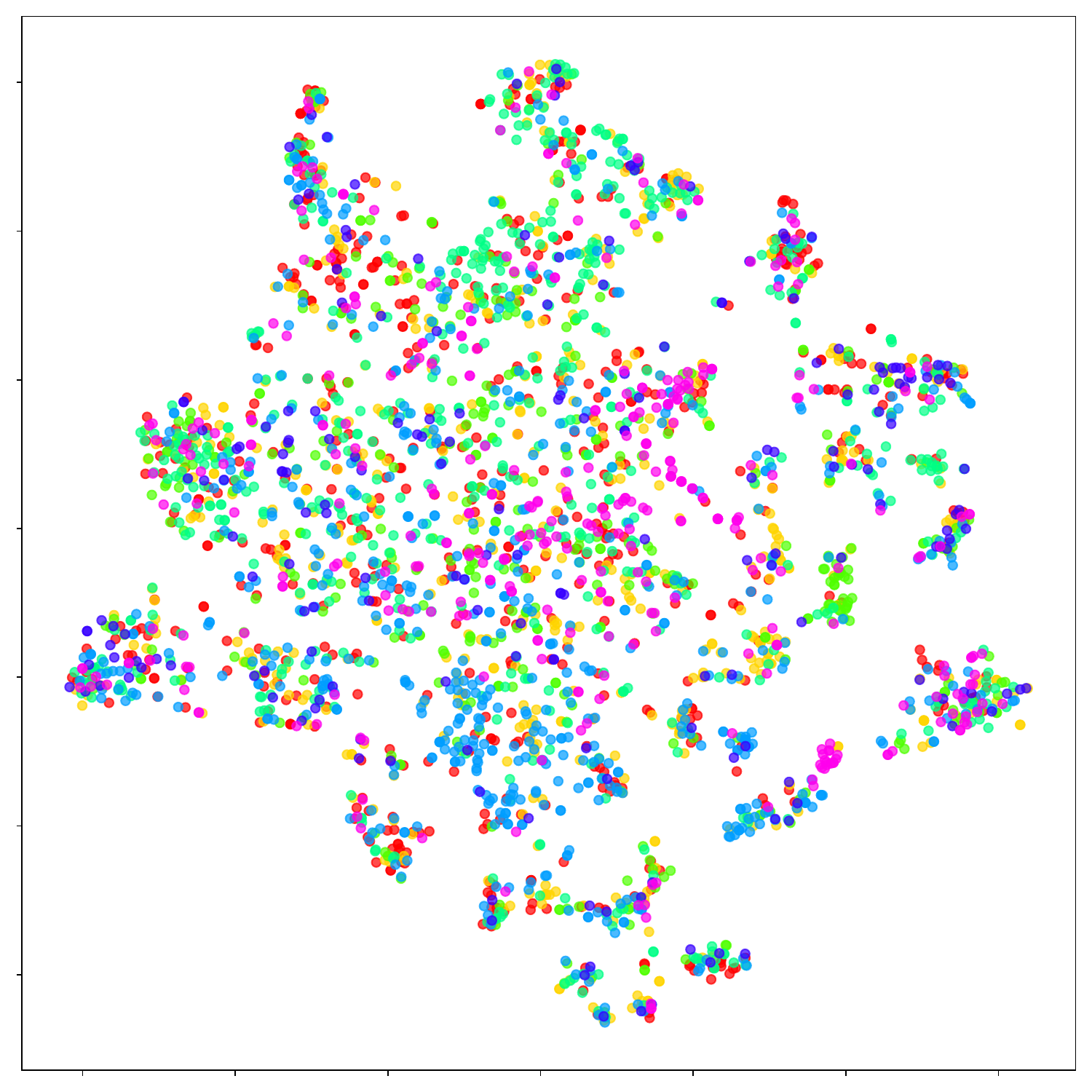}} &
         \subcaptionbox{True articulatory features + voicing\label{sfig:phoneme_representations-c}}{\includegraphics[width=0.225\linewidth]{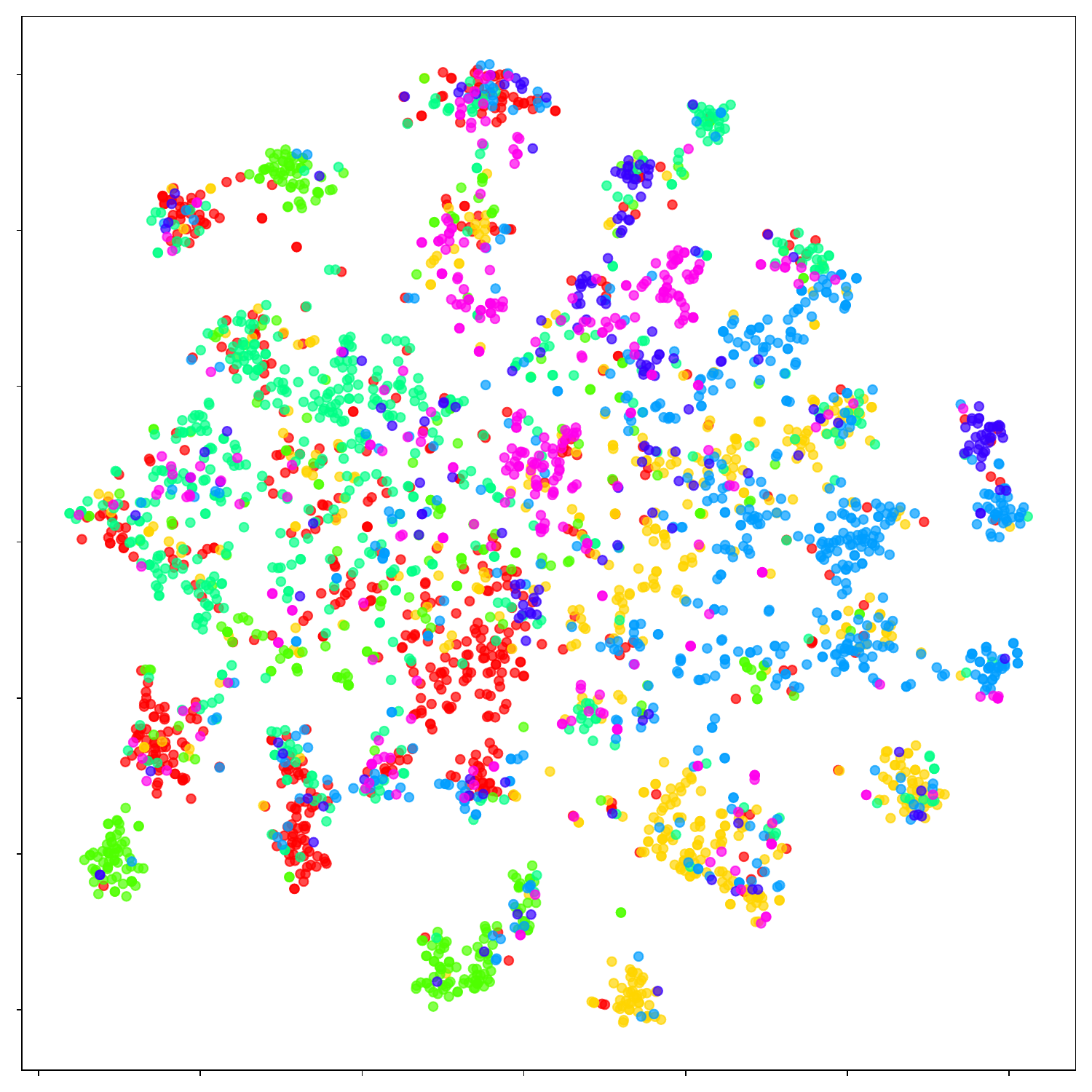}} &
         \subcaptionbox{Model-free articulatory features + voicing\label{sfig:phoneme_representations-d}}{\includegraphics[width=0.225\linewidth]{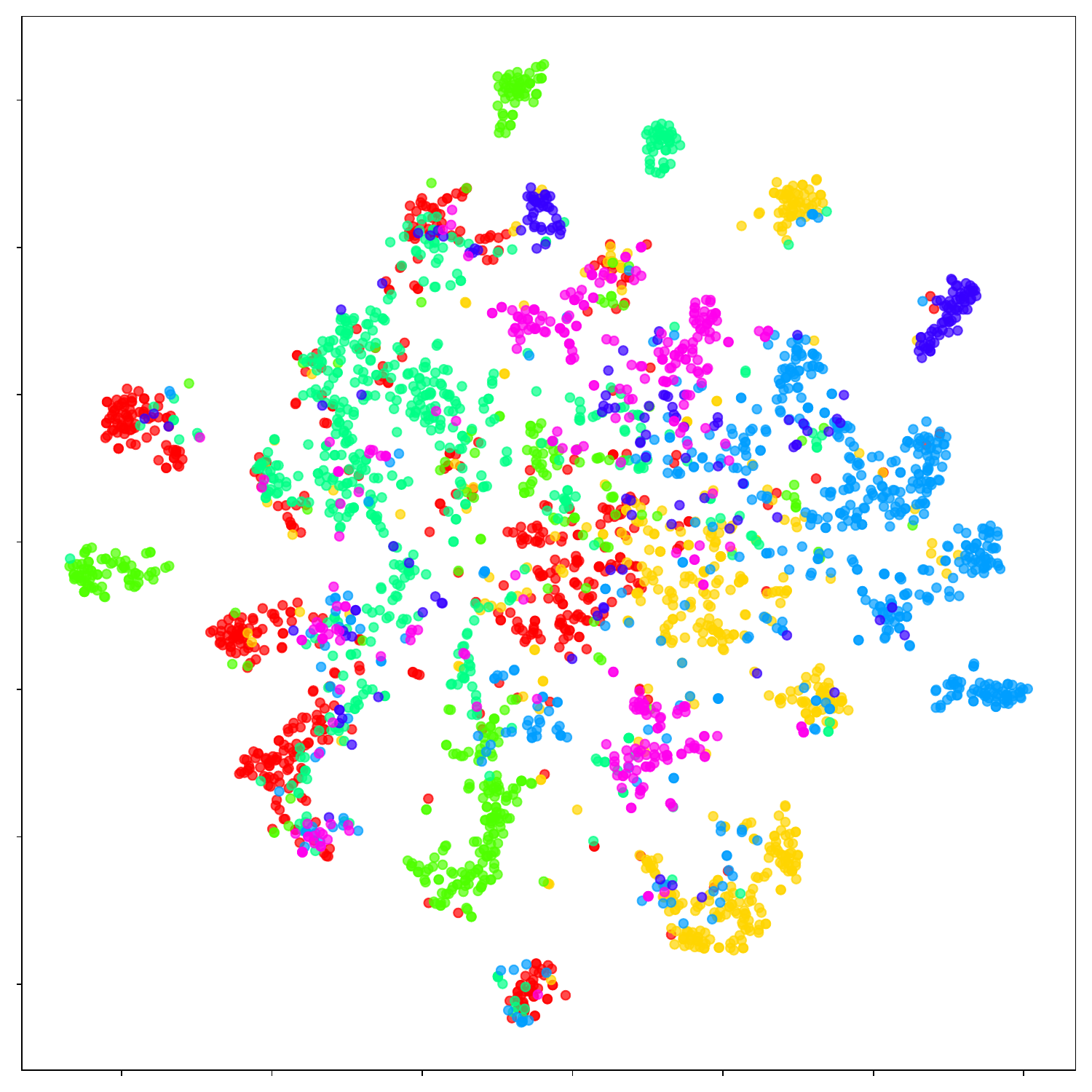}} &
         \subcaptionbox{Autoencoder-based articulatory features + voicing\label{sfig:phoneme_representations-e}}{\includegraphics[width=0.225\linewidth]{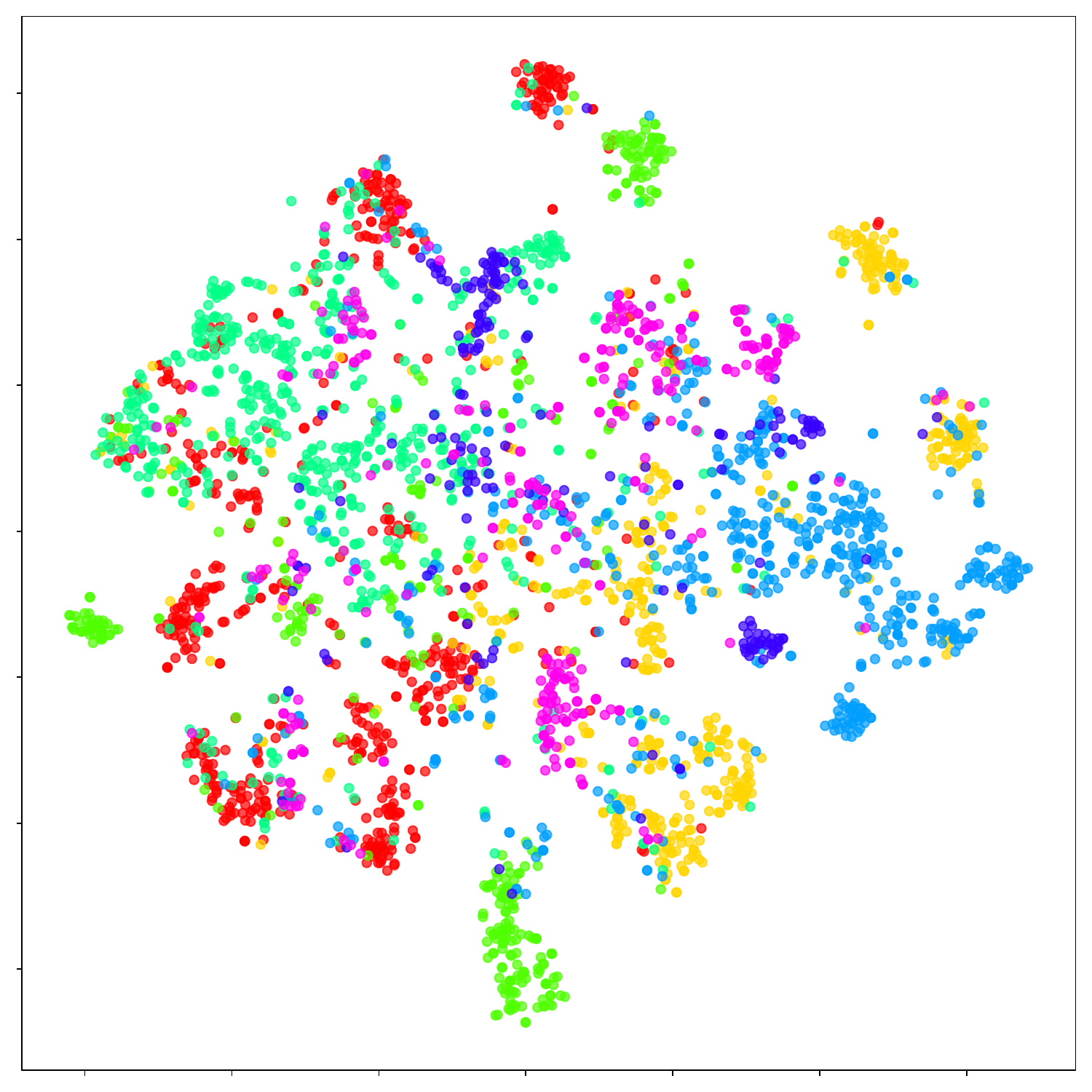}}\\
    \end{tabular}
    \caption[T-SNE representations of the recognized phonemes.]{T-SNE plot of the phoneme representations for each feature set. \textbf{Better visualized in digital form.}}
    \label{fig:phoneme_representations}
\end{figure*}

\autoref{tab:stats_per_model} presents the PER for each feature set. \autoref{fig:phoneme_representations} shows the t-SNE plots of the phoneme representations learned by each model. The phonemes were grouped into their respective phonetic classes in \autoref{fig:phoneme_representations} to facilitate reading and visualization, and it includes only the phonemes listed in \autoref{tab:phonetic_classes}.

\begin{figure*}[ht!]
    \centering
    \begin{tabular}{cccc}
    \subcaptionbox{Acoustic signal\label{sfig:substitution_matrix-a}}{\includegraphics[width=0.225\linewidth]{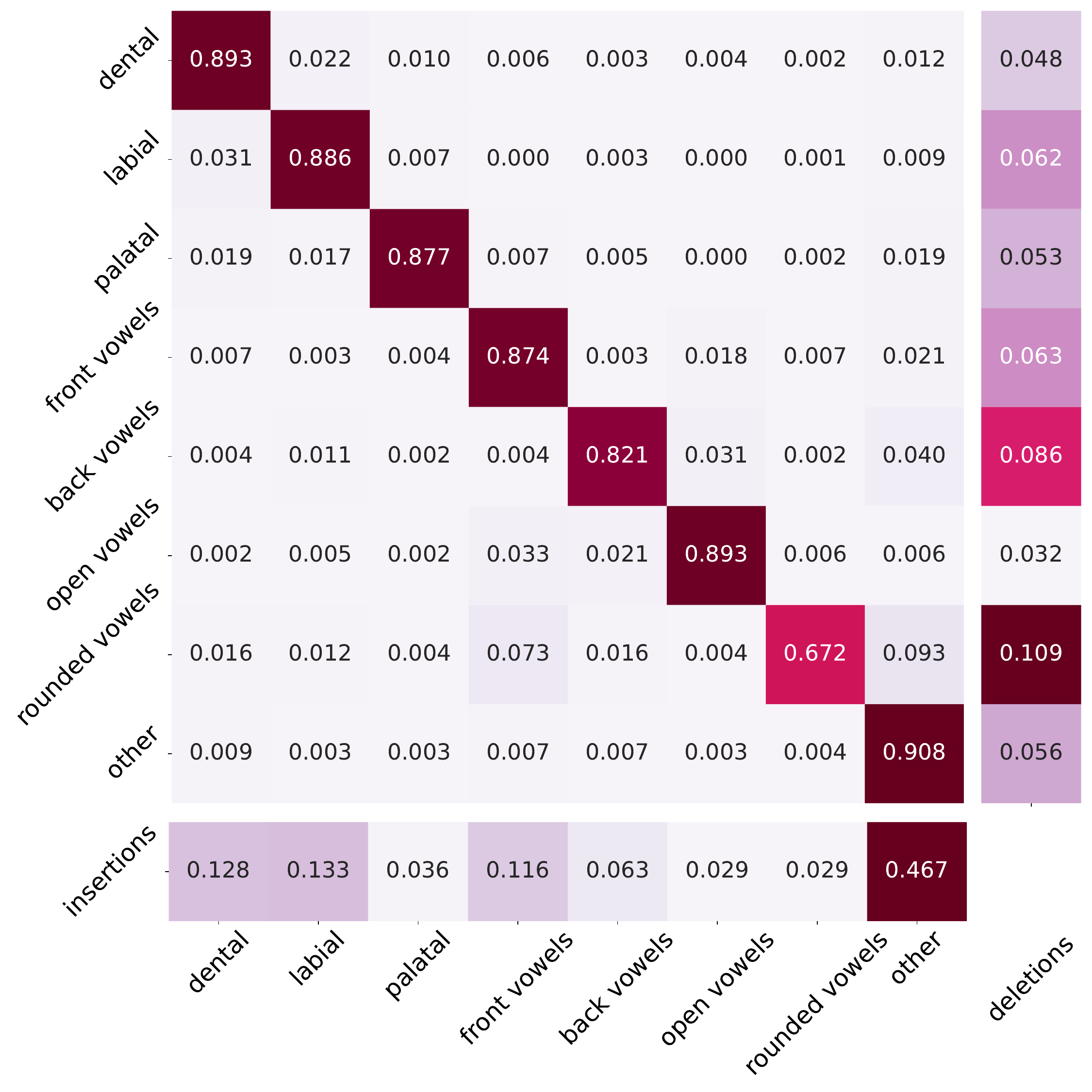}} &
    \subcaptionbox{True articulatory features + voicing\label{sfig:substitution_matrix-b}}{\includegraphics[width=0.225\linewidth]{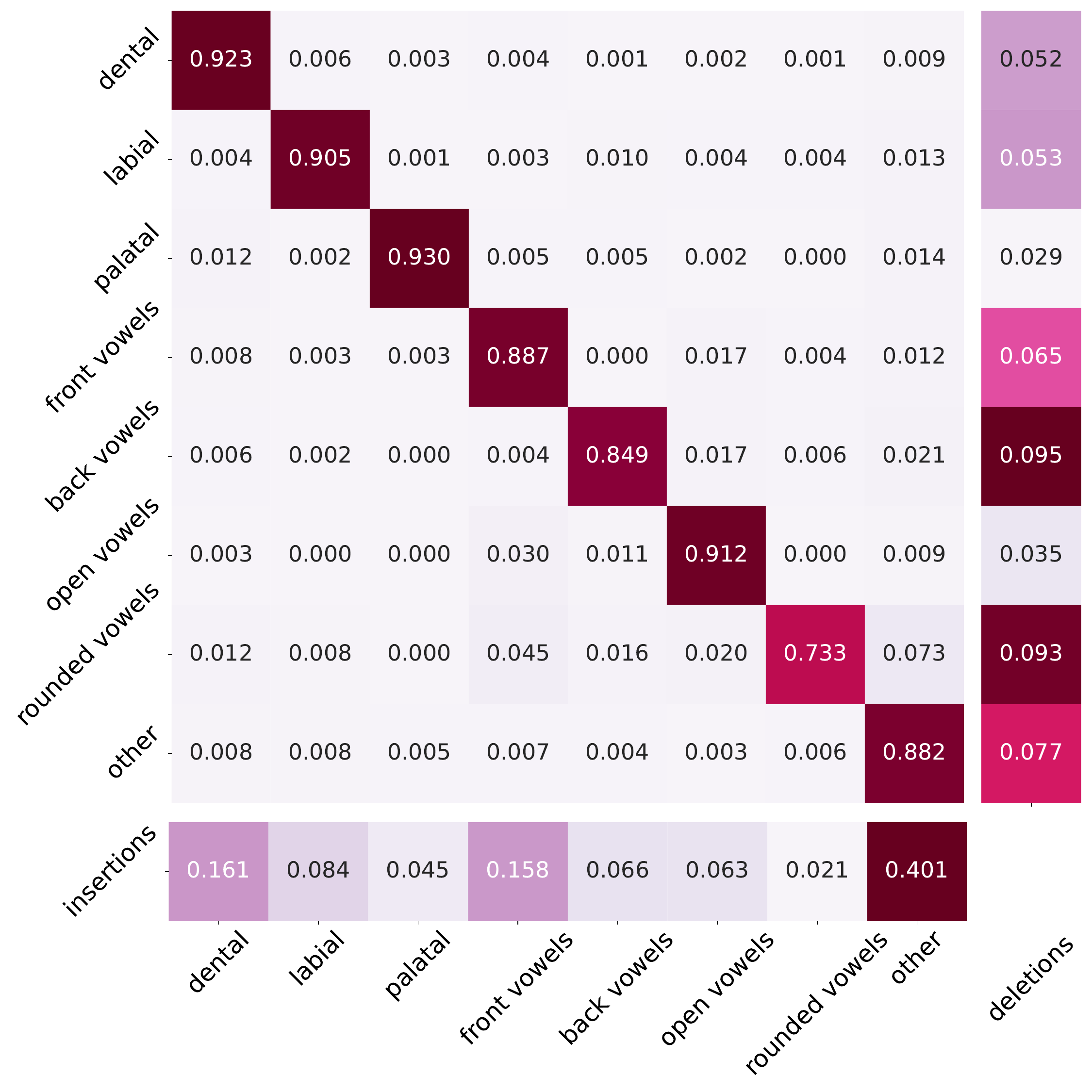}} &
    \subcaptionbox{Model-free articulatory features + voicing\label{sfig:substitution_matrix-c}}{\includegraphics[width=0.225\linewidth]{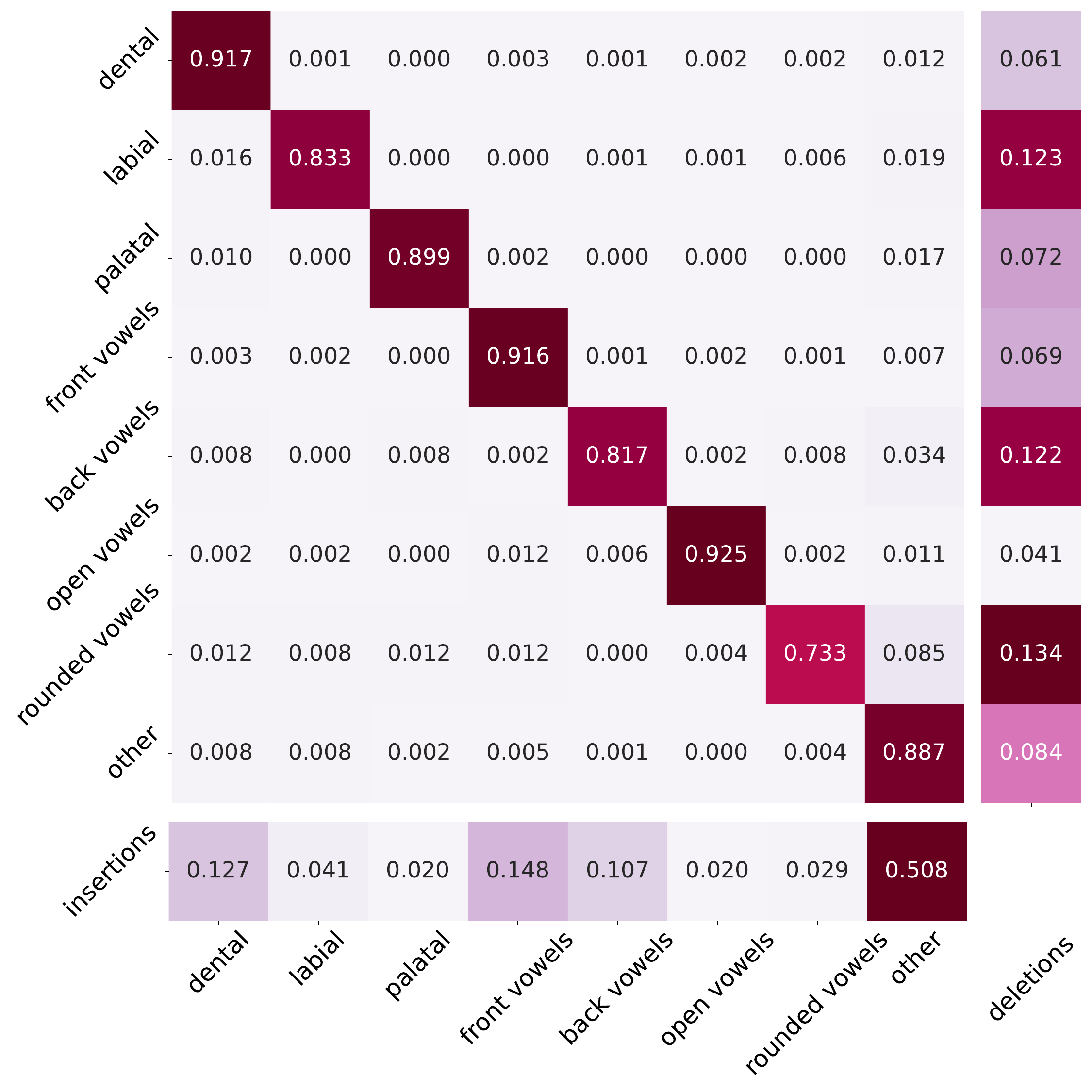}} &
    \subcaptionbox{Autoencoder-based articulatory features + voicing\label{sfig:substitution_matrix-d}}{\includegraphics[width=0.225\linewidth]{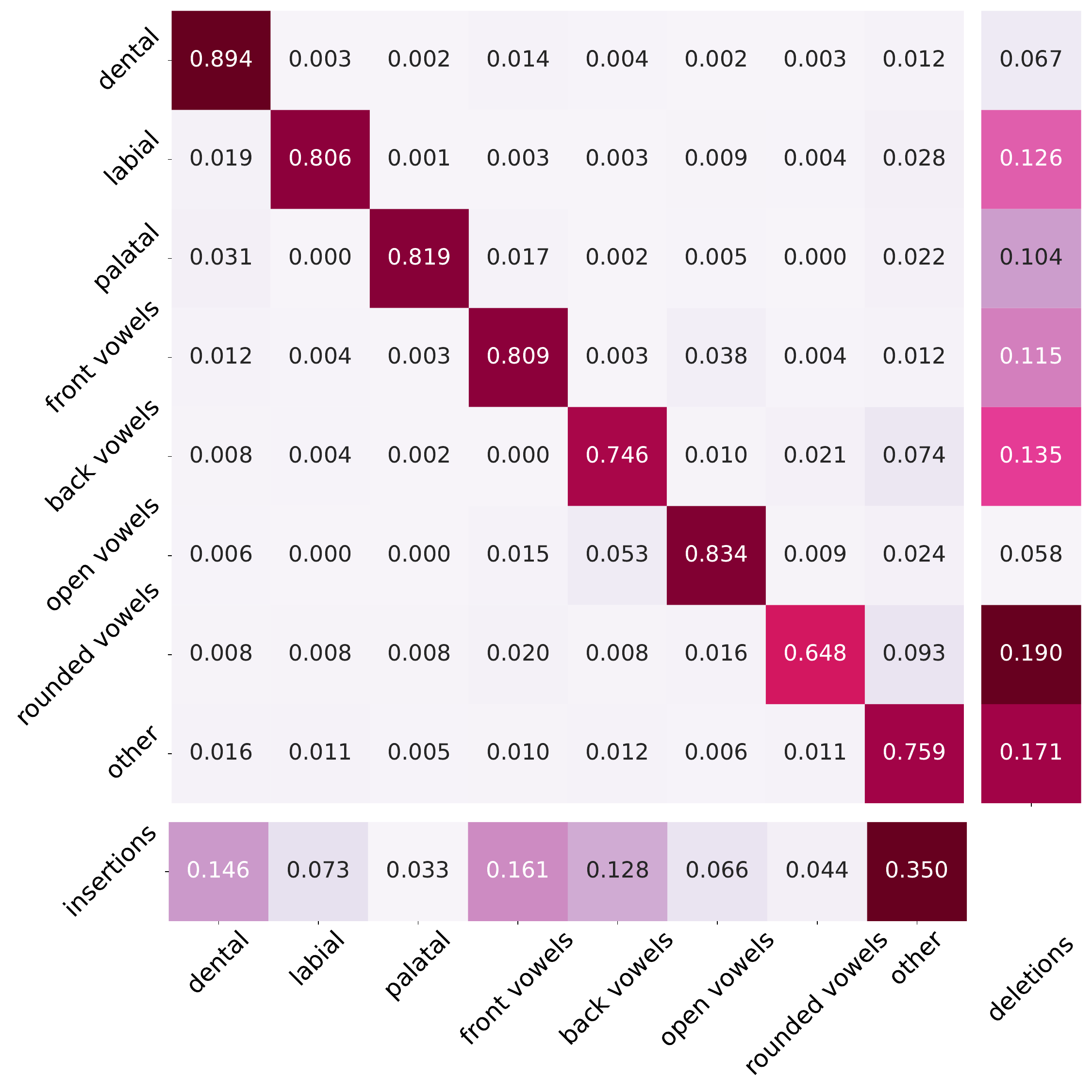}}
    \\
    \end{tabular}
    \caption[Phoneme recognition confusion matrix using the synthetic articulatory features.]{Phoneme recognition confusion matrix normalized by the true labels. \textbf{Better visualized in digital form.}}
    \label{fig:substitution_matrix}
\end{figure*}

\autoref{fig:substitution_matrix} displays the ASR confusion matrix of the phoneme recognition, with phonemes grouped into their phonetic classes. Similarly to the confusion matrix used for traditional classification tasks, the rows represent the actual classes, and the columns represent the predicted classes. Each cell $c_{ij}$ indicates the class $i$ being substituted by class $j$; hence the main diagonal represents correct matches. The last column represents the deletions of each class, while the last row represents the insertions of each class. It is important to highlight that since the matrix is normalized by the true labels, the deletions column displays different information than the insertions row. While the element $c_i$ in the deletions column shows the percentage of deleted tokens of class $i$, the element $c_j$ in the insertions row presents the percentage of insertions corresponding to class $j$.

\begin{table}
    \centering
    \caption[Phoneme error rate for the acoustic and articulatory features.]{PER for the acoustic and articulatory features, with and without voicing encoding.}
    \begin{tabular}{lcc}
        \toprule
        \thead{Feature Set} & \thead{Voicing\\Encoding} & \thead{PER} \\
        \midrule
        Acoustic Feat. & -- & 23.30 \\
        \midrule
        True Art. Feat. & No & 23.65 \\
        Phon.-Wise Mean-Contour Art. Feat. & No & 47.22 \\
        Model-Free Art. Feat. & No & 24.34 \\
        Autoencoder-Based Art. Feat. & No & 38.85  \\
        \midrule
        True Art. Feat. & Yes & 21.66 \\
        Phon.-Wise Mean-Contour Art. Feat. & Yes & 43.18 \\
        Model-Free Art. Feat. & Yes & \textbf{20.59} \\
        Autoencoder-Based Art. Feat. & Yes & 31.69  \\
        \bottomrule
    \end{tabular}
    \label{tab:stats_per_model}
\end{table}

\section{Discussion}

The comparison between our models and the state of the art requires attention. The main benchmark for the task is the TIMIT dataset~\cite{linguistic1990darpa}; wav2vec~\cite{schneider2019wav2vec} report a PER of 14.7 and wav2vec~2.0~\cite{baevski2020wav2vec} report a PER of 8.3 on it. However, these models are much larger than ours and trained with massive data. Additionally, our recorded audio contains an intense MRI noise and is damaged by the denoising algorithm, contrarily to TIMIT, which has clean speech. Nevertheless, most importantly, outperforming these models is not our goal.
Instead, we aim at quantifying the phonetic information retained by the articulatory features and the one reproduced by the vocal tract synthesizer.
These models are a reference for judging if the recognizer's predictions are \textit{good enough} to be used as a metric. That said, the models trained with the acoustic features and the articulatory features with voicing encoding resulted in a proper recognition compared to wav2vec but are still far from the results of wav2vec~2.0. Nevertheless, the results are satisfactory for our objective.

\autoref{tab:stats_per_model} shows that the recognition performance using the true articulatory features alone is indistinguishable from that of the acoustic features, which is a very satisfactory result since we expected that without the source information, the recognition would be much worse. Although surprising, the results are understandable. On the one hand, the articulators' contours extracted with the tracking method described in \cite{ribeiro2024automatic} are of high quality, showing outstanding performance in a multi-speaker setting. Despite the higher error in contact regions, the overall quality compensates for the errors. On the other hand, the substantial MRI noise in the acoustic features and the deterioration due to denoising contribute to a lower performance with the acoustic features.

Even if the articulatory features alone present performance very close to the acoustic signal, it is hard to believe that it retains the complete phonetic information. The vocal tract shapes lack source information, meaning unvoiced phonemes are indistinguishable from their voiced counterparts. After adding voicing encoding to the feature set, performance improved by $1.99$ points.

Unsurprisingly, the phoneme-wise mean contour presents inferior recognition performance, which is expected due to the model's simplicity, which does not account for contextual information.
The PER using synthetic vocal tract shapes from the model-free approach with source information is outstanding. The recognition performance has a lower PER than all other feature sets, including the true articulatory features with voicing encoding, even if the latter corresponds to the same features used during training. Even if the model-free articulatory features are of high quality and the vocal tract shapes are realistic, the result is surprising. The reason might be that the articulatory synthesizer filters out noise in the true features, generating cleaner speech articulations.
Conversely, recognition performance using the autoencoder-based articulatory features is lower, only beating the mean-contour features.
Even if the model-free and autoencoder-based systems presented very competitive results so far, we see that the PER can discriminate the two models more meaningfully. Phoneme recognition captures our initial impression that the model-free system yields speech articulations with higher temporal consistency.

\autoref{fig:phoneme_representations} shows that by adding voicing encoding the articulatory features form apparent groups in the embedding space that are not seen even with the acoustic features even if the recognition is not included in the synthesizers' optimization procedure. The PER and the feature embeddings corroborate the quality of synthesized vocal tract shapes.

We need to address the issue of reaching the correct places of articulation. \cite{ribeiro2023deep} discussed the difficulty of achieving proper dental, palatal, and labial constrictions with the model-free system. It should not be a surprise that the model has a high deletion rate for phonemes that require the lips to be correctly approximated, such as labials, back and rounded vowels (\autoref{sfig:substitution_matrix-c}), which is not observed with the true articulatory (\autoref{sfig:substitution_matrix-b}) and acoustic features (\autoref{sfig:substitution_matrix-a}). Another factor that helps explaining high deletion rates for rounded vowels is the lack of lip rounding in the data.


The confusion matrix for the autoencoder-based system (\autoref{sfig:substitution_matrix-d}) retains high deletion rates for dental, labial, and palatal phonemes even though we observed an improvement in these places of articulation. Since deletion rates with autoencoder-based articulatory features are higher for all phonetic classes, it is unclear whether high deletion rates for these specific classes are due to poor recognition performance or lack of proper articulatory constrictions.

\section{Conclusion}

This paper explored phoneme recognition for evaluating speech articulation synthesis conditioned on the sequence of phonemes to be articulated. The metrics available so far were objective, but captured only some of the desired dimensions. Metrics like point-to-closest-point distance penalize models that account for variability in articulation, which limits their ability to handle intra-speaker variability and, consequently, multi-speaker scenarios. Alternatively, tract variables measure the dynamics of speech and the interaction between articulators but may not adequately represent vowels. Thus, this paper focuses on developing an evaluation system that encourages models to synthesize an intelligible articulatory feature set, independent of the speaker, while fitting a broader phonetic context.

We evaluated three articulation synthesizers from the literature. The first model, serving as a baseline, is the average vocal tract shape for each phoneme. The second is a model-free vocal tract synthesizer, which directly generates vocal tract shapes from the sequence of phonemes to be articulated without relying on an articulatory model. The third is an autoencoder-based system that generates articulatory parameters of an articulatory model of speech, designed using an autoencoder.

Thorough evaluation of these models indicates that the autoencoder-based approach aligns more closely with articulatory phonology literature~\cite{browman1992articulatory} by producing more accurate places of articulation. However, subjective evaluation suggests that the model-free system generates more temporally stable articulations. In summary, previous research highlights that evaluating and comparing speech articulation synthesis is more complex than initially thought.

To address this challenge, we trained a phoneme recognizer capable of transcribing articulatory features into phonemes. Our experiments showed that the true articulatory features extracted from RT-MRI are phonetically rich and contain even more information than the acoustic baseline. This finding aligns with our laboratory experience: the recorded audio data are often compromised by MRI noise and the limitations of the denoising algorithm. For example, the vocal tract contours were more recognizable than some audio fragments for trained professionals in phonetics and vocal tract acoustics.

The most important result is that ASR can be used to assess the relevance and efficiency of dynamic vocal tract shape generation approaches. Indeed, the evaluation provided by automatic recognition is in line with the degree of phonetic refinement we have introduced into the synthesis model, and provides an objective measure of our qualitative perception of the results.

Here, our research focuses on a single speaker due to practical constraints. Recording speech data with RT-MRI is costly, and health restrictions limit participant availability. 
However, extending this research to multi-speaker settings is a crucial next step. Since phoneme recognition is inherently speaker-independent, a promising research direction is investigating how it could be integrated into the training pipeline of speech articulation synthesis models to achieve implicit speaker normalization.

\bibliographystyle{IEEEtran}
\bibliography{mybib}

\end{document}